\documentclass[10pt,fleqn,a4paper,twoside]{article}
\usepackage{abcm}
\usepackage{caption}
\usepackage{subcaption}
\usepackage{amsmath}
\usepackage{tikz}
\usetikzlibrary{decorations.pathmorphing,patterns}
\usetikzlibrary{shapes,arrows}
\usepackage{siunitx}

\def\shortauthor{E. Vergamini, L. Santos, C. Zanette, Y. Moreno, F. Escalante and T. Boaventura}
\def\shorttitle{Construction of an Impedance Control Test Bench}

\begin{document}
\fphead
\hspace*{-2.5mm}\begin{tabular}{||p{\textwidth}}
\begin{center}
\vspace{-4mm}
\title{COB-2023-0847\\ %{\color{red}(XXXX is the identification number of the final paper)}
CONSTRUCTION OF AN IMPEDANCE CONTROL TEST BENCH}
\end{center}

\authors{Elisa Gamper Vergamini} \\
\institution{elisa.vergamini@usp.br} \\

\authors{Leonardo Felipe dos Santos} \\
\institution{leonardo.felipe.santos@usp.br} \\

\authors{Cícero L. A. Zanette} \\
\institution{cicero\_zanette@usp.br} \\

\authors{Yecid Moreno} \\
\institution{yecidmoreno@usp.br} \\

\authors{Felix M. Escalante} \\
\institution{felix.escalante@unesp.br} \\

\authors{Thiago Boaventura} \\
\institution{tboaventura@sc.usp.br} \\

\institution{Universidade de São Paulo - Escola de Engenharia de São Carlos, Av. Trabalhador Sancarlense, 400, 13565-090, São Carlos/SP, Brazil} \\ %(If all authors are from the same institution, the "Institution and address" must be placed only once.)

\institution{São Paulo State University - Institute of Science and Technology Sorocaba, Av. Três de Março, 511,  18087-180, Sorocaba/SP, Brazil.
 } \\
%\institution{e-mails} \\

%\institution{e-mail} \\
\\
%\authors{Same format for other authors, if any} \\
\\
\abstract{\textbf{Abstract.} Controlling the physical interaction with the environment or objects, as humans do, is a shared requirement across different types of robots. To effectively control this interaction, it is necessary to control the power delivered to the load, that is, the interaction force \emph{and} the interaction velocity. However, it is not possible to control these two quantities independently at the same time. An alternative is to control the relation between them, with Impedance and Admittance control, for example. The Impedance Control 2 Dimensions (IC2D) bench is a test bench designed to allow the performance analysis of different actuators and controllers at the joint level. Therefore, it was designed to be as versatile as possible, to allow the combination of linear and/or rotational motions, to use electric and/or hydraulic actuators, with loads known and defined by the user. The bench adheres to a set of requirements defined by the demands of the research group, to be a reliable, backlash-free mechatronic system to validate system dynamics models and controller designs, as well as a valuable experimental setup for benchmarking electric and hydraulic actuators. This article presents the mechanical, electrical, and hydraulic configurations used to ensure the robustness and reliability of the test bench. Benches similar to this one are commonly found in robotics laboratories around the world. However, the IC2D stands out for its versatility and reliability, as well as for supporting hydraulic and electric actuators.}
\\
\\
\\
\end{tabular}

\section{INTRODUCTION}

It is increasing the deployment of robots in situations that are more difficult to navigate, such as interacting with people or dealing with unpredictable and unfamiliar terrain. In light of this scenario, robot-environment interaction management is a challenge. Force and velocity define the physical interaction, so impedance control, which can control the relation between these magnitudes, has been the focal point ever since \citep{Hogan1984}. However, it is worth noting that such controllers always have a force control loop \citep{calanca2017}. In the case of manipulator robots, quadrupeds, and bipeds, the selected actuators functioned as the robot's muscles. That is to say, in the development process of a new robot, the actuator selection is a valuable step to ensure that power, inertia, and task-related requirements are covered. Therefore, there needs to be a way to understand and compare the performance of different actuators associated with various controllers.

The design of joint level test benches for these benchmarks, such as ForceCAST, may consist of a motor module, a spring module (for SEA), a torque sensor module, and a virtual environment/load module \citep{vicario2021benchmarking}. The latter is a counter motor or a lever that can be excited manually. However, it only supports small rotational electric actuators (around \SIrange{0}{10}{\newton\meter}). Another bench developed for the same purpose is the Metrox, which, like the ForceCAST, has two motors connected, with a torquemeter between them. This one supports electric motors with larger sizes and is limited to rotational cases tough \citep{Bahler21}.
Another bench used for the same purpose and was the top reference for this project was the FC2D (Force Control 2 Directions), in which linear hydraulic and electric actuators are used, with the same structure as one motor connected to the other \citep{Hammer16}. Although this bench supports electric and hydraulic actuators, it does not support rotational actuators.

Considering the existing benches, the Impedance Control 2 Dimensions (IC2D) is an upgraded FC2D. The main goal was to build a modular bench capable of supporting electrical and hydraulic actuators, both rotational and linear, to render various environments virtually (via software) or actually. In this paper, we will describe the mechatronic construction of the bench, starting with the mechanical structure and the requirements accomplished, passing through the selection of materials and manufacturing, the actuators and sensors used, the design of the hydraulic circuit, and the electronic circuit. Finally, we present the bench validation based on the repeatability and accuracy of the results obtained.

%%%%%%%%%%%%%%%%%%%%%%%%%%%%%%%%%%%%%%%%%%%%%%%%%%%%%%%%%%%%%%
\section{MECHANICAL STRUCTURE}

\begin{figure}[h!]
    \centering
    \includegraphics[width =0.8\columnwidth]{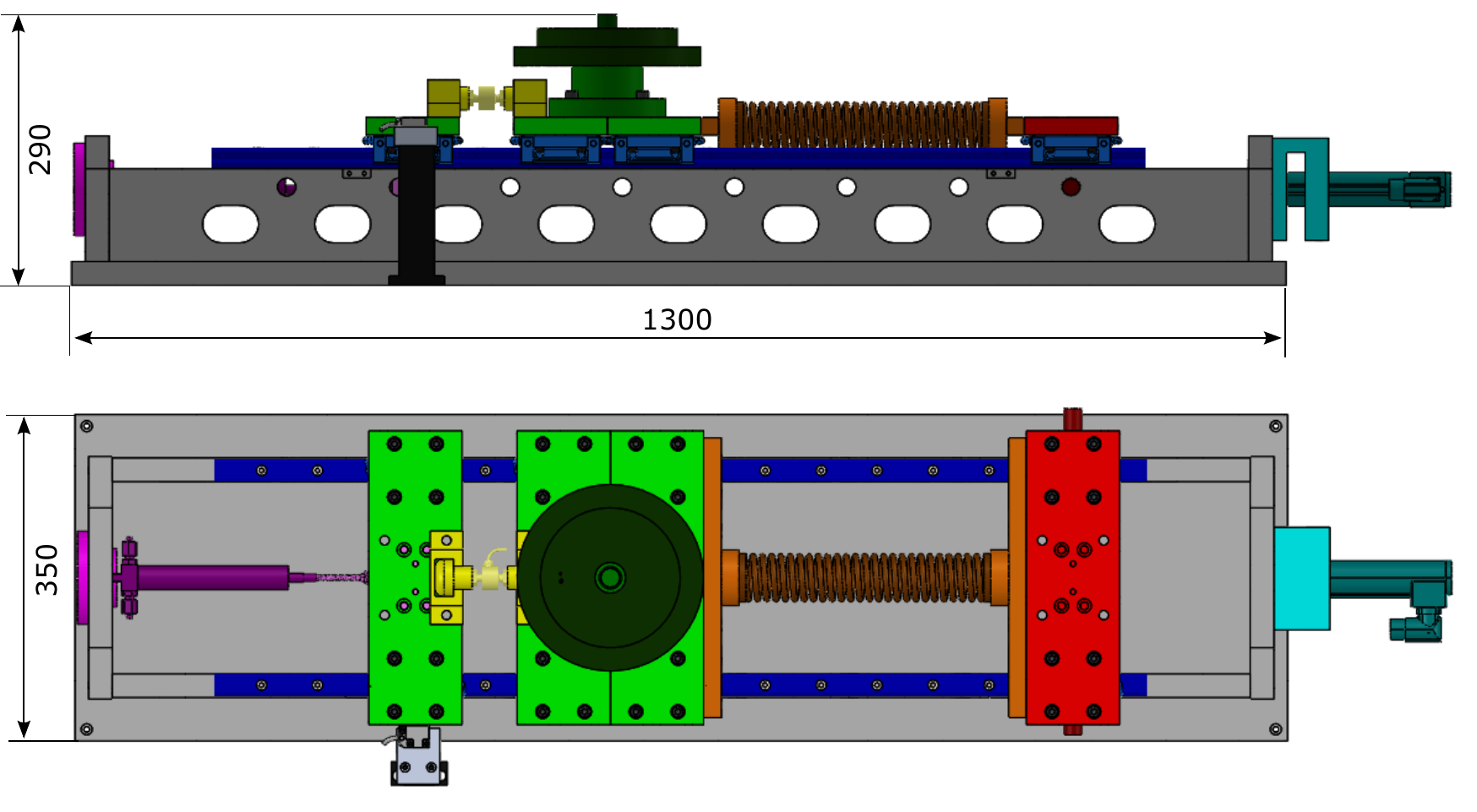}
    \caption{Bench final assembly, its main dimensions, in mm. The parts represented in green are those that are able to move; in dark blue are the sliders and guides; in red are the parts that will be blocked; in purple and cyan are the actuation systems; in yellow the force sensor; and in black on the side the position sensor.}
    \label{fig:global}
\end{figure}

Initially, the structure of the bench and the main mechanical and performance requirements will be explained. The IC2D test bench is a platform with two colinear degrees of freedom (DoFs). It aims at the study and comparison of actuation systems and controllers at the joint level. The structure consists of linear guides, rollers, and cars. It can test different hydraulic and/or electric actuators, linear and/or rotational. It also has a coupling system for load cells, springs, and dampers. The design is modular, i.e., the parts are versatile and interchangeable, making it easy and quick to change actuators, couplings, sensors, and setups. One of the possible setups is in Fig.~\ref{fig:global}. 

The project aimed at, in its most complex assembly, the need for two cars with variable inertia and two cars to transmit the actuator force. It also aimed to have easily removable springs and shock absorbers, to allow the coupling of encoders and force sensors in general, to easily change the actuators, to be able to block the actuator or one of the ends of a physical environment and to have two actuators, not necessarily of the same model, simultaneously attached, so that the second is able to cause disturbances in the system, to eventually, simulate different environment configurations. The maximum mass of 10 kg was stipulated for each slider and the nominal force, discounting peaks, is expected to be around 5 kN (linear). For the rotational case this value will be lower. The bench must be robust, respecting the necessary strength for the tests, and it should be fixed on a breadboard, to damp mechanical vibrations. The main components of the workbench are described in more detail below.

\subsection{Translations cars}

\begin{figure}[htb]
    \centering
        \begin{subfigure}[t]{0.45\textwidth}
            \centering
            \includegraphics[width = 1\columnwidth]{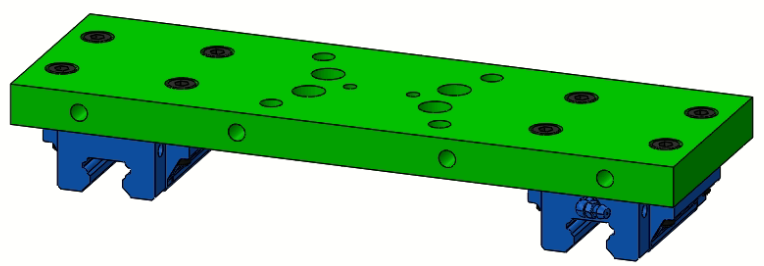}
            \caption{Standart platform}
            \label{fig:std_plataform}
        \end{subfigure}
        \hfill
        \begin{subfigure}[t]{0.45\columnwidth}
            \centering
            \includegraphics[width = 1\columnwidth]{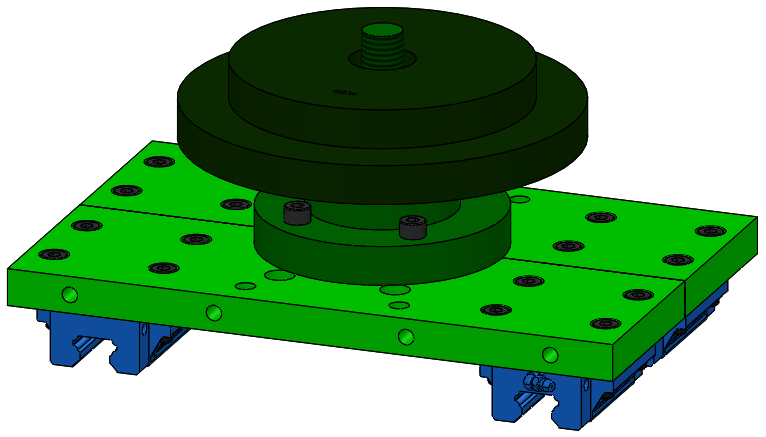}
            \caption{Load car}
            \label{fig:load_car}
        \end{subfigure}
        \hfill
        \begin{subfigure}[t]{0.45\columnwidth}
            \centering
            \includegraphics[width = 1\columnwidth]{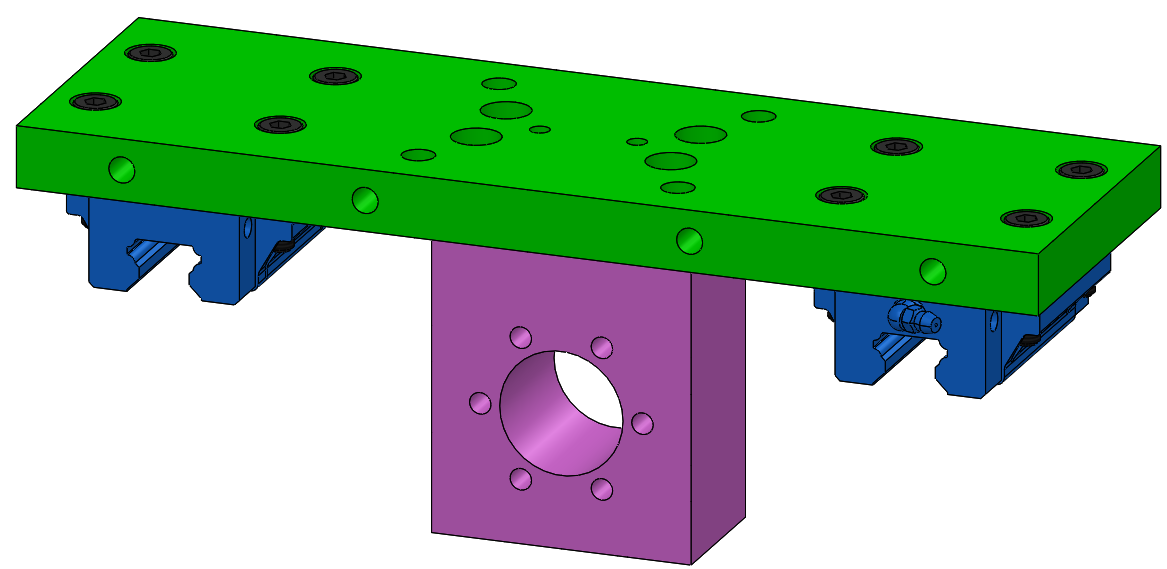}
            \caption{Transmission car}
            \label{fig:transmission_car}
        \end{subfigure}
        \hfill
        \begin{subfigure}[t]{0.45\columnwidth}
            \centering
            \includegraphics[width = 1\columnwidth]{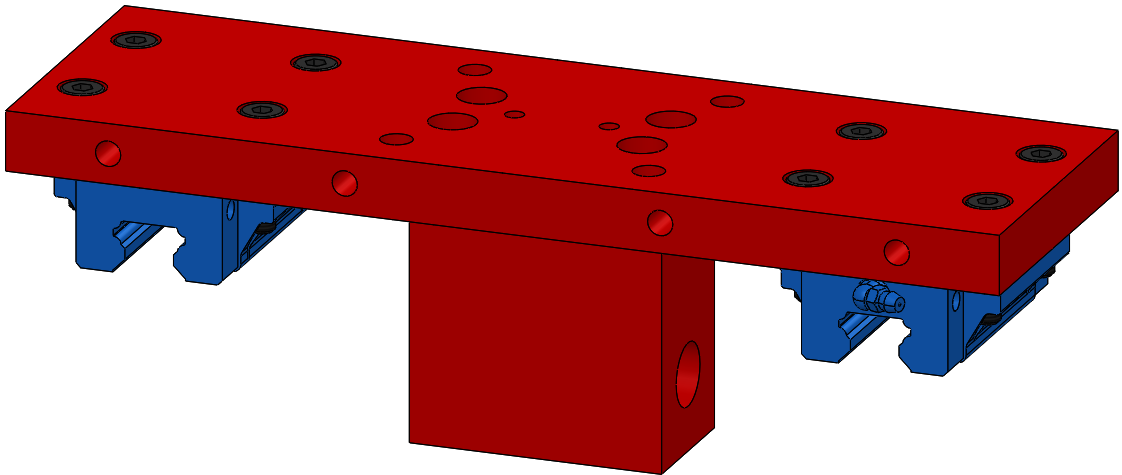}
            \caption{Blockage car}
            \label{fig:blocked_car}
        \end{subfigure}
        \vspace{0.3cm}
    \caption{The parts represented in light green are those that are able to move; in dark blue are the sliders; in red are the parts that will be blocked; in purple is the block for the transmission of force; and in dark green is the union part, that joins two platforms and supports variable weights.}
    \label{fig:car_config}
\end{figure}

To simplify manufacturing, the mobile part of the bench is made up of a standard platform attached to two sliders, which can be changed according to the additional parts selected. There are three types of translation cars: the load car (Fig.~\ref{fig:load_car}), the transmission car (Fig.~\ref{fig:transmission_car}), and the blockage car (Fig.~\ref{fig:blocked_car}), and it is possible to use the standard platform as a connection car if necessary.
The standard platform is the one in Fig.~\ref{fig:std_plataform}, it is used for the three configurations mentioned above. 

\subsubsection{Transmission car}

\begin{figure}[htb]
    \centering
        \begin{subfigure}[b]{0.45\columnwidth}
            \centering
            \includegraphics[width =1\columnwidth]{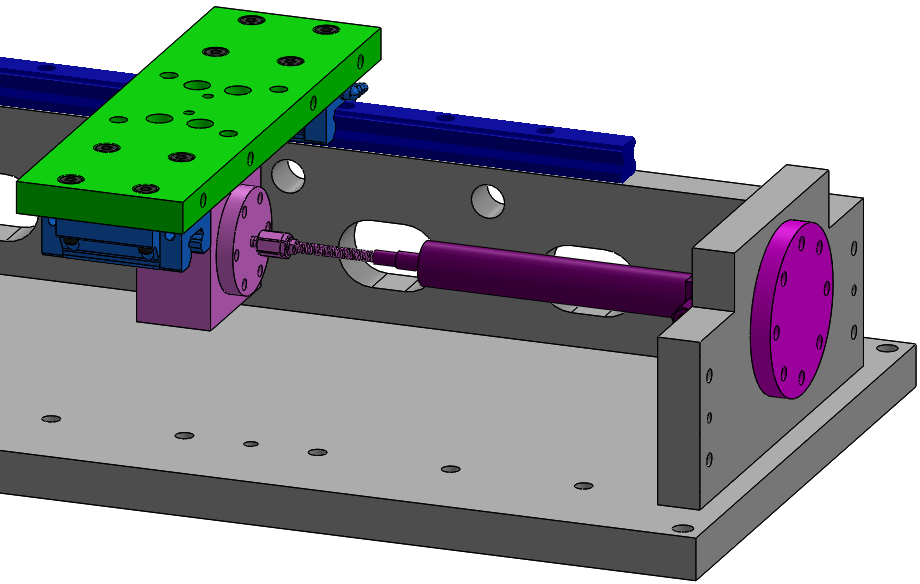}
            \caption{In purple we have highlighted the hydraulic linear actuation system, in which we highlight the flange for fixing the cylinder.}
            \label{fig:cilindro}
        \end{subfigure}
        \hfill
        \begin{subfigure}[b]{0.45\columnwidth}
            \centering
            \includegraphics[width =1\columnwidth]{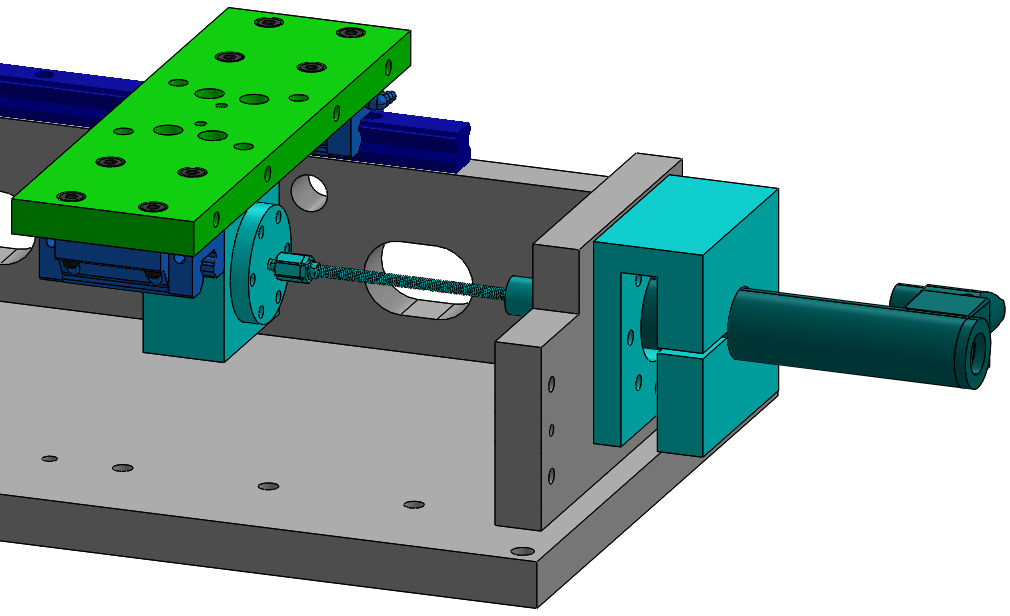}
            \caption{In cyan we have highlighted the electric linear actuation system, in which we highlight the flange for fixing the LinMot.}
            \label{fig:eletrico}
        \end{subfigure}
        \vspace{0.3cm}
    \caption{Linear transmission assembly: In gray you have the global base, the sides and the flanges to connect the actuators, in light green the standard platform and in dark blue the linear guides and sliders.}
    \label{fig:linear_transmission}
\end{figure}

The platform is attached to a hybrid block, that is, it can be assembled with a nut/sleeve with a spindle or a flange with an angular offset compensation, the first for rotational actuators and the second for linear actuators. For the linear transmission car, there are currently two possible configurations: the hydraulic (Fig.~\ref{fig:cilindro}) or electric actuation system (Fig.~\ref{fig:eletrico}). Both are fixed to the transmission block by the same angular and radial offset compensation. However, given the distances between the actuators, their stroke, and the desired assembly setup, it is necessary to insert spacers to extend the actuator rods. For this purpose, the threaded bars were cut to the required dimensions to ensure assembly without backlash. 

For the rotational transmission car, analogously to the linear transmission one, the same block was used, and a socket was designed to brace a sleeve/nut with a spindle, capable of converting the rotational movement of the actuator into linear for the car. The universal ball sleeve/nut initially selected is the BN25x5R NOWPR and the spindle is the BN5x5R 90 G5, both from \textit{SKF}.

\subsubsection{Load car}

To assemble the load car, the union part represented in Fig.~\ref{fig:load_car} connects two modular platforms into one. On the same part, it is possible to attach a threaded bar, which will serve as a guide for the iron weight plates. The height of the union part is assigned to avoid collisions of the plates with the load cell coupling assembly and also to evenly distribute the load between the four sliders that support the whole set.

\subsubsection{Blockage car}

The blockage car was developed to allow the use of only one actuator with a known load, as well as to characterize the actuator from a test with it completely blocked, both tests normally used in benchmarking methodologies. For this purpose, a 20~mm diameter steel shaft functions as a latch, precisely dimensioned to withstand the force distributed by the contact surface of the shaft with a block of up to 5~kN. The shaft has an interference fit with the block and structure's side supports, so it has no backlash.

\subsection{Structure, actuators flanges, linear guide and sliders}

The linear guides are fixed on supports that are attached to a global base, at the end of the guides there are two universal flanges that allow the passage of both the cylinder shaft and the spindle axis. The flanges have a hole pattern, which allows the manufacture of adjustable fixing parts for each actuator to be attached to it, as shown in Fig. \ref{fig:linear_transmission}. The selected linear guide has a length of one meter in a singular part. To allow more fluid movement of the sliders, the selected model is the TRH25-FL from \textit{Kalatec}. Their static and dynamic load capacity exceeded the desired 10 kN. It is possible to fit up to six sliders per guide. The choice of linear guides provided a gap between the supports for the guides and the platforms, a space that can be useful for instrumentation wiring. When assembling the linear guides, it was necessary to use a dial gauge to ensure that the geometric tolerances defined in the manufacturer's datasheet were respected. Furthermore, to eliminate friction in the system as much as possible, a lubricating oil (SAE10) was applied.

\subsection{Load Cell and linear encoder}

\begin{figure}[htb]
    \centering
        \begin{subfigure}[b]{0.30\textwidth}
            \centering
            \includegraphics[width =1\columnwidth]{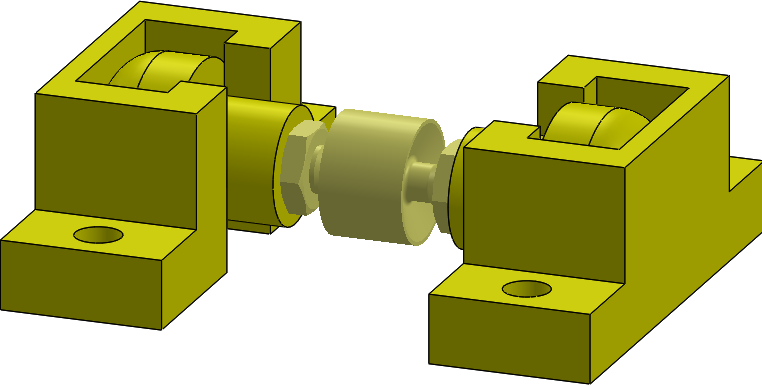}
            \caption{}
            \label{fig:loadcell1}
        \end{subfigure}
        \hfill
        \begin{subfigure}[b]{0.30\columnwidth}
            \centering
            \includegraphics[width =1\columnwidth]{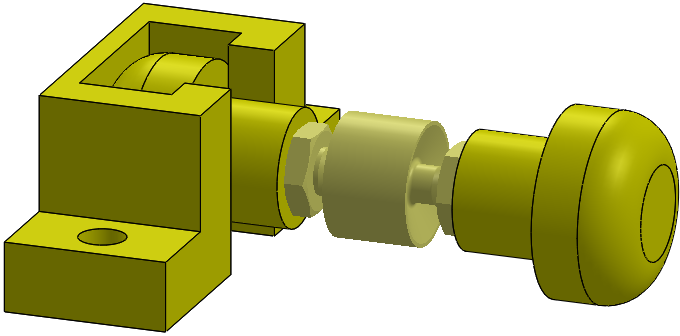}
            \caption{}
            \label{fig:loadcell2}
        \end{subfigure}
        \hfill
        \begin{subfigure}[b]{0.30\columnwidth}
            \centering
            \includegraphics[width =1\columnwidth]{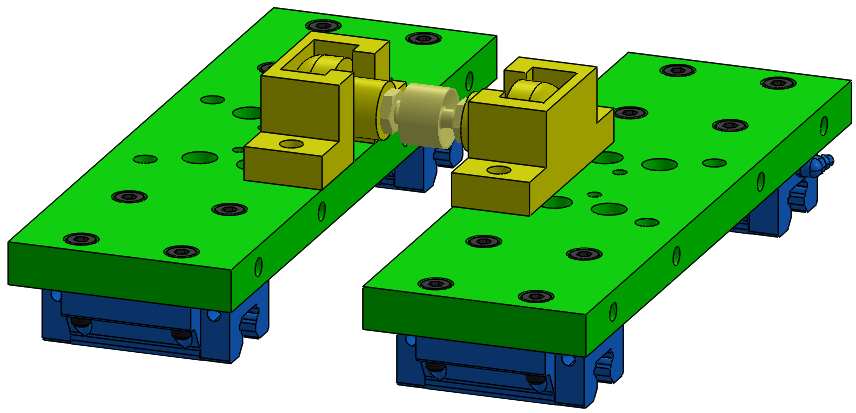}
            \caption{}
            \label{fig:loadcell3}
        \end{subfigure}
        \vspace{0.3cm}
    \caption{In yellow the entire load cell fastening set is represented. The cell represented is the Burster 8417-6005.}
    \label{fig:loadcell}
\end{figure}

The fixation of the load cell assembly has two main parts (Fig.~\ref{fig:loadcell1}), the first is responsible for attaching the load cell to the platform, the second is threaded and manufactured specifically for each sensor. The curved surface of the former part compensates for few angular misalignments to avoid leaving residual stresses in the reading of the sensor, however, granting no backlash in the assembly (Fig.~\ref{fig:loadcell2}). Once attached to the coupling, the load cell is then fixed between two platforms, as shown in Fig.~\ref{fig:loadcell3}.

\subsection{Springs and dampers fixation}

\begin{figure}[htb]
    \centering
    \includegraphics[width =0.5\columnwidth]{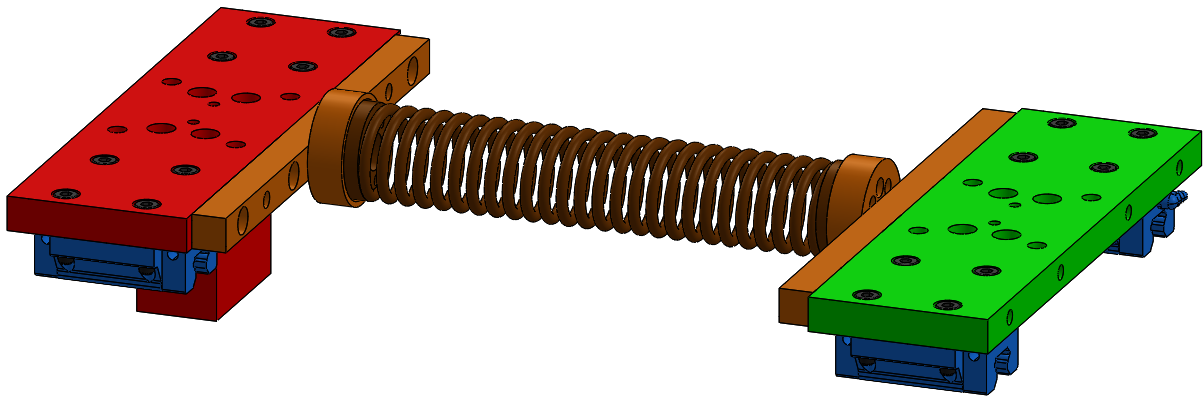}    
    \vspace{0.3cm}
    \caption{Spring assembly currently used is represented in orange.}
    \label{fig:mola_montagem}
\end{figure}

The fixation of springs and dampers on the platforms was designed to ensure the symmetry of the system. These combinations allow the implementation of different environments and even the use of series elastic actuators (SEA). Currently, the springs used are those adapted from another bench, but there are plans to acquire a greater variety of springs and dampers. Therefore, four holes were drilled on each side of the platforms, allowing the coupling of different types of springs, dampers, or even supplementary parts, as shown in Fig.~\ref{fig:mola_montagem}. In order to use the springs already available, a flange and an additional part that joins the flange to the side of the platform were made.

\subsection{Material selection}

\begin{figure}[htb]
    \centering
    \includegraphics[width =0.8\columnwidth]{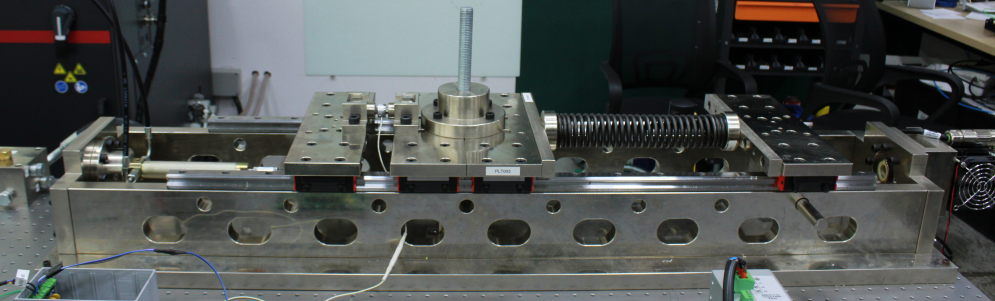}
    \caption{IC2D bench final assembly}
    \label{fig:fotoglobal}
\end{figure}

All the structural parts, platforms, and coupling parts were initially proposed in the aluminum alloy 6082 to give the desired strength and a total mass that is not so high compared to any simple carbon steel. However, since the bench would have a modular characteristic, that is, it is going to be assembled and disassembled frequently, it was decided to use SAE 1020 steel, as the threads needed in the assembly could be worn out with use if aluminum was used. Consequently, the final mass of the bench can reach up to 200 kg depending on the assembly configuration. However, since the bench is on a breadboard, it is not necessary to continue moving the base structure (set with greater mass contribution). As steel can oxidize with time, it was necessary to perform a surface treatment to preserve the parts, galvanic nickel was chosen, which would leave a more refined surface layer compared to zinc plating, in addition to having a more pleasant aesthetic, as shown in Fig.~\ref{fig:fotoglobal}.

\subsection{Good Design Practices}

All parts were designed to have simplified manufacturing, in addition to allowing all geometric tolerances to be guaranteed at the time of assembly. For example, the use of guide pins both to guarantee positioning and to withstand shear forces on the screws. An alignment nail was placed on one of the sides to facilitate the alignment of the rails and the smooth sliding of the sliders. All CAD designs, technical drawings and assemblies are available in \cite{IC2D2023}. 

\section{FLUID POWER SYSTEM} \label{sec:fluid_power_design}

\begin{figure}[htb]
    \centering
        \begin{subfigure}[b]{0.474\columnwidth}
            \centering
            \includegraphics[width =1\columnwidth]{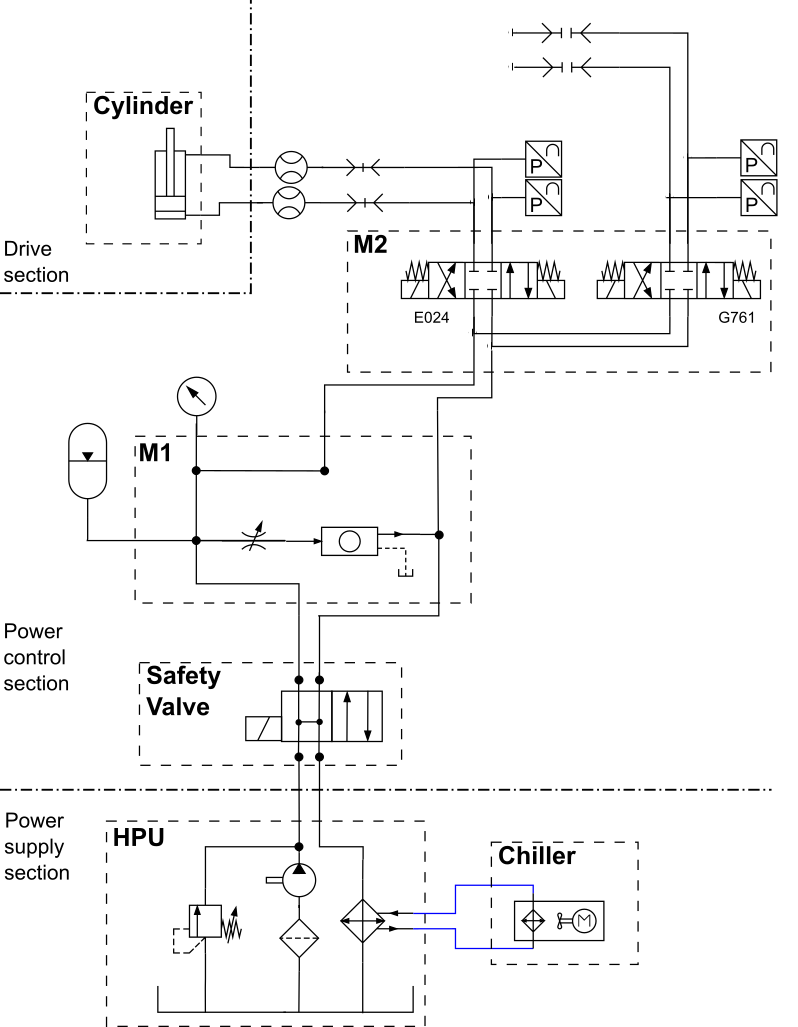}
            \caption{Hydraulic circuit diagram.}
            \label{fig:hydraulic_diagram}
        \end{subfigure}
        \hfill
        \begin{subfigure}[b]{0.450\columnwidth}
            \centering
            \includegraphics[width =1\columnwidth]{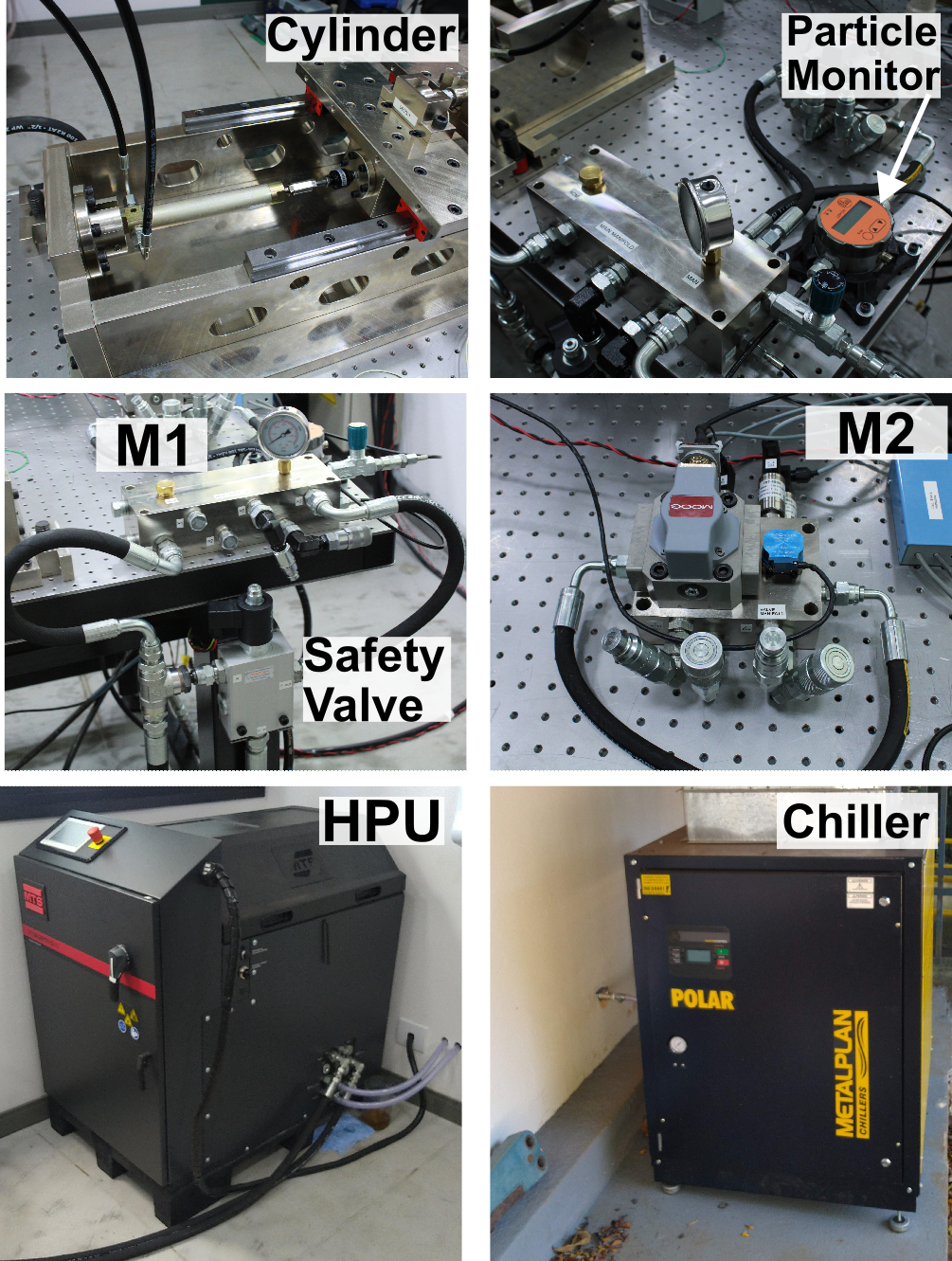}
            \caption{Fluid power hardware.}
            \label{fig:hydraulic_hardware}
        \end{subfigure}
    \vspace{0.3cm}
    \caption{Fluid power system overview containing the main elements of the power supply, power control, and drive sections. The accumulator and flow sensors are shown only in the diagram (a). The hydraulic circuit was inspired by \cite{seminiThesis}.}
    \label{fig:hydraulic_system}
\end{figure}

The fluid power system was designed considering the requirements for the hydraulic actuator on the IC2D and other experimental platforms at the Legged Robotics Group facility, such as the hydraulic robotic leg and hydraulic arm manipulator \citep{semini2008hyq, ur2016design}. The key elements were selected or designed according to technical requirements to guarantee performance and reliability. An overview of the hydraulic circuit diagram and the relevant hardware is shown in Fig.~\ref{fig:hydraulic_system}. A detailed description of these elements is presented in the following sections.

\subsection{Hydraulic Power Unit} \label{sec:fluid_power_hpu}

The Hydraulic Power Unit (HPU) is a variable displacement pump from MTS, model HPU 515-11 Water Cooled, with regulated operating pressure up to \SI{20.7}{\mega\pascal}, and \SI{1.67e-4}{\cubic\meter\per\second} flow. Filtration is in full flow on the return side and an \SI{18.5}{\kilo\watt} electric motor drives the pump shaft. The HPU contains pressure, oil level, and temperature sensors. A heat exchanger with a heat removal capacity of \SI{18.7}{\kilo\watt} is also in the HPU. Furthermore, the oil reservoir capacity is about \SI{0.2}{\cubic\meter}. The pressure and flow supported by the HPU meet the hydraulic requirements to supply the IC2D servovalves.

\subsection{Chiller} \label{sec:fluid_power_chiller}

The chiller is required to supply a coolant fluid to the HPU. Temperature control is one of the main aspects of fluid power systems. According to the standard, the oil temperature must be \SI{40}{\degreeCelsius} at work. Temperature oscillations are not desired to preserve oil integrity and viscosity stability. The HPU heats the oil in the reservoir through the operation of its own motor that turns the pump. However, it is necessary to remove excessive heat, so the chiller supplies cooled water in the range of \SI{5}{\degreeCelsius} up to \SI{25}{\degreeCelsius}. The installed chiller model is the PA-15 RE, from Metalplan, with a heat removal capacity of \SI{17.4}{\kilo\watt}.

\subsection{Hydraulic hoses and connections} 

The hoses connect the fluid system and allow setup flexibility, a drawback of tubes compared to hoses. In this setup, hoses were preferred, with a working pressure of \SI{21.5}{\mega\pascal}, above the maximum working pressure supported by the HPU. The hose diameter and length specifications considered pressure loss and assembly rules in accordance with DIN 20021 standards. The connections standard adopted was BSP and DKO for better sealing. Quick connector couplings are present for easy switching in some points of the circuit.

\subsection{Manifolds} 

Multiple fluid lines are necessary due to the different hardware fed by the HPU. To distribute these lines, a distribution manifold was designed and manufactured. Moreover, a servovalves manifold was also designed. The distribution manifold (M1) was designed with the capability of six different lines, support for a particle counter line, pressure gauge and accumulator (Fig.~\ref{fig:manifold1_assembly}). The servovalves manifold (M2) was designed to match the port pattern of the Moog's servovalves E024 and G-761, and with support for pressure sensors on the actuation lines of both valves (Fig.~\ref{fig:manifold2_assembly}). The main manifold was manufactured in SAE~1020 carbon steel and the valve manifold in 6061 aluminum alloy, and to prevent rust, the steel manifold was surface treated as the bench parts.

\subsection{Hydraulic oil} 

Oil is the element through which mechanical power transmission occurs. Mineral oil is the most common fluid adopted in robotics. Its low compressibility makes energy transfer faster and more effective. In addition, oil lubricates circuit components such as valves and cylinders, preserving their integrity over time and reducing leakage. The fluid selected for this system is the ISO~68 mineral hydraulic oil, with \SI{875}{\kilo\gram\per\cubic\meter} of density and standard kinematic viscosity of \SI{6.8e-5}{\square\meter\per\second} at \SI{40}{\degreeCelsius}. The fluid Bulk module at atmospheric pressure and without mixed gas is equivalent to \SI{1.34}{\giga\pascal}, with a compressibility rate of 0.75~$\%$ per \SI{10}{\mega\pascal}. The fluid class is HLP according to the DIN~51524-2 standard, characterized by increased corrosion resistance, wear protection, and aging stability.

\subsection{Auxiliary elements} 

In addition to the above elements, auxiliary elements such as a pressure gauge, an accumulator, and a particle counter are required to support better system efficiency and maintainability. The pressure gauge is installed on M1, allowing the pressure check of the system. An accumulator will be installed to prevent pressure oscillations in the system. The particle counter, or contamination sensor, is installed on M1. The sensor model is LDP100, from IFM Electronic, able to monitor the oil particle level in compliance with ISO~406~/~NAS 1638. This sensor is critical to ensure that both valves, E024 and G-761, work with increased lifespan, preventing clogging.

\subsection{Servovalves and cylinder}

\begin{figure}[htb]
    \centering
        \begin{subfigure}[b]{0.3\columnwidth}
            \centering
            \includegraphics[width =0.7\columnwidth]{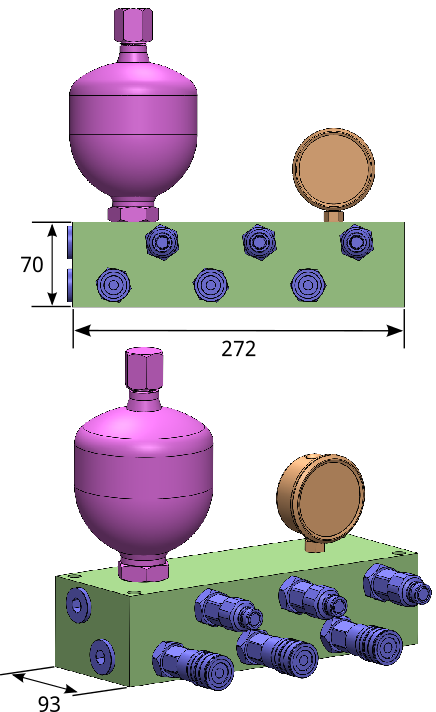}
            % [width =0.7\columnwidth]
            \caption{Distribution manifold assembly.}
            \label{fig:manifold1_assembly}
        \end{subfigure}
        %\hfill
        \begin{subfigure}[b]{0.3\columnwidth}
            \centering
            \includegraphics[width =1\columnwidth]{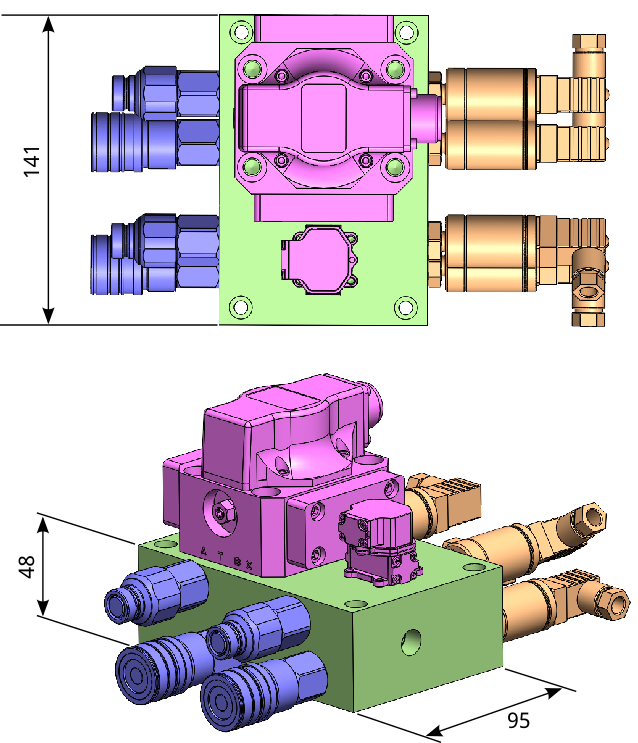}
            \caption{Servovalves manifold assembly.}
            \label{fig:manifold2_assembly}
        \end{subfigure}
        \vspace{0.3cm}
    \caption{Fluid power system manifolds assemblies for functionality preview. The manifolds are green, and the quick connectors are blue on both figures. In (a), the accumulator and the pressure gauge are pink and orange, respectively. In (b), the servovalves and the pressure sensors are pink and orange, respectively. Manifold sizing handled mechanical strength and line displacement specifications, providing the required space to attach multiple connectors and sensor parts shown. The drawing dimensions are in millimeters.}
    \label{fig:manifolds_assembly}
\end{figure}

In perspective to assess the force and impedance control on hydraulic actuators used on quadruped robots, the selected cylinder is the model LB6 1610 0080, double acting type from Hoerbiger, and one of the servovalves is the E024 four-way directional, from MOOG. These are the same components used in the hydraulic quadruped HyQ \citep{semini2011design, fahmi2020stance}. The cylinder has a stroke of \SI{80}{\milli\meter}. The diameters of the pistons are \SI{16}{\milli\meter} and \SI{10}{\milli\meter}. The E024 servovalve bandwidth is \SI{250}{\hertz} at \SI{21}{\mega\pascal} and \SI{1.25e-4}{\cubic\meter\per\second} maximum rated flow at \SI{7}{\mega\pascal} pressure drop. A second servovalve, model G-761 with \SI{450}{\hertz} bandwidth, also from MOOG, is mounted in the servovalves manifold (Fig.~\ref{fig:manifold2_assembly}) for control system characterization with a wider servovalve bandwidth.

%%%%%%%%%%%%%%%%%%%%%%%%%%%%%%%%%%%%%%%%%%%%%%%%%%%%%%%%%%%%%% 

\section{ELECTRONICS}\label{sec:electronic_design}

Data acquisition and control commands were processed using the microcontroller STM32 Nucleo-F446RE, from STMicroelectronics, with an ARM32-bit Cortex-M4 CPU, real-time capability up to \SI{180}{\mega\hertz}, and \SI{512}{\kilo\byte} flash memory. Since this MCU uses a \SI{3.3}{\volt} signal level, an electronic I/O signal conversion for actuators and sensors was made using a customized board with the following capabilities: \SI{10}{\volt} and \SI{5}{\volt} to \SI{3.3}{\volt} conversion for the sensors, \SI{3.3}{\volt} to \SI{-10}{\volt}/\SI{10}{\volt} for the actuators and an amplifier circuit for the load cell. The full schematic of the custom board can be seen in \cite{IC2D2023}.

Regarding the actuators, both drivers have the same signal input, i.e. \SIrange{-10}{10}{\volt}. The linear motor used was the PS01-37x120F-HP-C (stator) and PL01-20x240/180-HP (slider), from LinMot, with \SI{255}{\newton} of maximum force and \SI{120}{\milli\meter} of maximum stroke. The power supply for the linear motor was the S01-72/1000. The driver used for the LinMot was the B1100-GP-HC. The driver used for the servo valve was the G123-815A001, from MOOG. The load force was measured using the SMT1-250 load cell from Interface, ranging \SI{250}{\newton}, nominal sensitivity of \SI{2}{\milli\volt\per\volt}, for the electric tests, and a miniature load cell 8417-6005 from Burster, ranging \SI{5}{\kilo\newton}, nominal sensitivity of \SI{1}{\milli\volt\per\volt}, for the hydraulic tests. The pressure in the control chambers was measured with the pressure sensors NAT 8251.74.2517, from Trafag AG, with \SI{25}{\mega\pascal} maximum pressure reading and $\pm$ 0.5~$\%$ accuracy. The incremental encoder used was the LM10IC001AB10F00 from RLS, with a resolution of \SI{1}{\micro\meter}. The software implemented on the microcontroller is the ForceCAST Studio, provided by Altair Robotics Laboratory \citep{vicario2021benchmarking}. The functional diagram of the electronics is shown in Figure~\ref{fig:electronics}.

\tikzstyle{block} = [draw, rectangle,  minimum height=10mm, minimum width=17mm, node distance=24mm]
\tikzstyle{point} = [coordinate, node distance=25mm]
\tikzstyle{pinstyle} = [pin edge={to-,thin,black}]

\begin{figure}[hbt]
    \centering
        \begin{tikzpicture}
        %We start by placing the blocks
            \node[block, align=center] (PC) {Computer};
            \node[block, right of=PC, align=center] (STM) {Nucleo\\F446RE};
            \node[point, right of=STM] (aux1) {};
            \node[block, above of=aux1, align=center, node distance=15mm] (Board1) {Custom\\Board};
            \node[block, below of=aux1, align=center, node distance=15mm] (Board2) {Custom\\Board};
            \node[block, right of=Board1, align=center] (driverV) {Valve\\Driver};
            \node[block, right of=Board2, align=center] (driverLM) {LinMot\\Driver};
            \node[block, right of=driverV] (Valve) {Valve};
            \node[block, right of=driverLM] (LM) {LinMot};
            \node[point, right of=Valve] (aux2) {};
            \node[block, below of=aux2, node distance=15mm] (sys) {System};
            \node[block, right of=sys] (enc) {Encoder};
            \node[block, above of=enc, align=center, node distance=15mm] (PS) {Pressure\\Sensor};
            \node[block, below of=enc, align=center, node distance=15mm] (LC) {Load\\Cell};
            \node[point, below of=driverLM, node distance=10mm] (aux3) {};
            \node[point, right of=enc, node distance=14mm] (aux4) {};
            \node[point, below of=aux3, node distance=5mm] (aux5) {};
            \node[point, above of=driverV, node distance=10mm] (aux6) {};

            % % % % Once the nodes are placed, connecting them is easy. 
            \draw[<->] (PC) -- (STM);
            \draw[<->] ([yshift=5mm]STM) -| (Board1);
            \draw[<->] ([yshift=-5mm]STM) -| (Board2);
            \draw[->] (Board1) -- (driverV);
            \draw[->] (driverV) -- (Valve);
            \draw[->] (Board2) -- (driverLM);
            \draw[->] (driverLM) -- (LM);
            \draw[<->] (Valve) |- ([yshift=5mm]sys);
            \draw[<->] (LM) |- ([yshift=-5mm]sys);
            \draw[->] (sys) |- (PS);
            \draw[->] (sys) -- (enc);
            \draw[->] (sys) |- (LC);
            \draw[->] (LC) |- (aux3) -| (Board2);
            \draw[->] (enc) -- (aux4) |- (aux5) -| (STM);
            \draw[->] (PS) |- (aux6) -| (Board1);   
        \end{tikzpicture}
        \vspace{0.3cm}
    \caption{Functional diagram of the electronic connections, each actuator must have your own custom board.}
    \label{fig:electronics}
\end{figure}
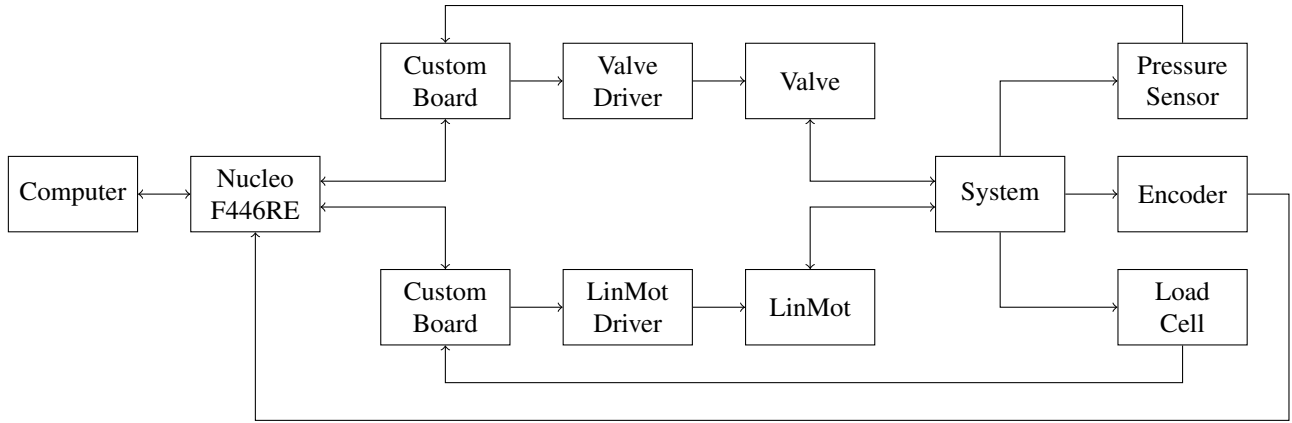

%%%%%%%%%%%%%%%%%%%%%%%%%%%%%%%%%%%%%%%%%%%%%%%%%%%%%%%%%%%%%%
\section{EXPERIMENTAL VALIDATION}

We verified three aspects to validate the construction of the bench. The first aspect is the experimental repeatability, or how closely are the measured data along multiple runs under the same setup. The second is the assessment of backlash-related phenomena measuring the displacement of a blocked car under the force of an actuator. Finally, the third is the assessment of friction forces on the bench as well as the calibration of the sensors. Knowing the stiffness of the spring used in the setup, we blocked one end of the spring and, using force and position sensors calculated the experimental stiffness of the spring.

\begin{figure}[htb]
    \centering
        \begin{subfigure}[b]{0.45\columnwidth}
            \centering
            \includegraphics[width =1\columnwidth]{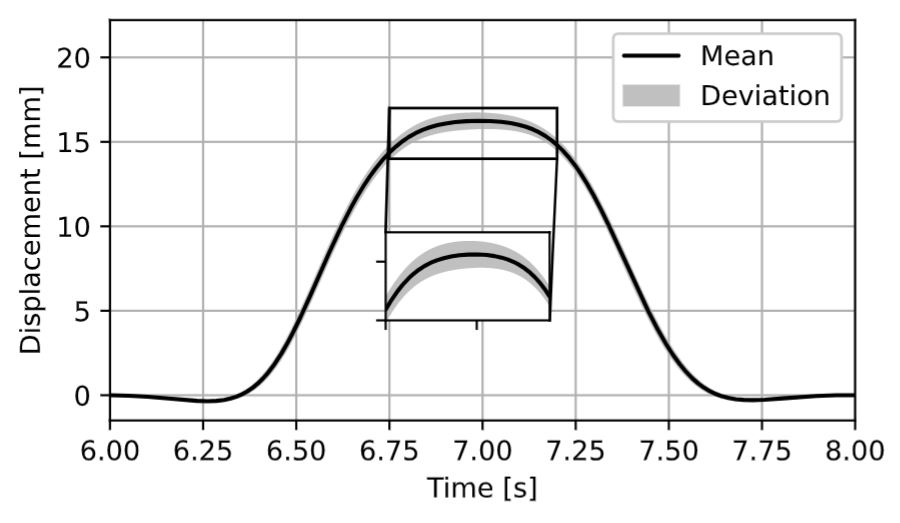}
            % [width =0.7\columnwidth]
            \caption{Displacement}
            \label{fig:hyd_disp}
        \end{subfigure}
        %\hfill
        \begin{subfigure}[b]{0.45\columnwidth}
            \centering
            \includegraphics[width =1\columnwidth]{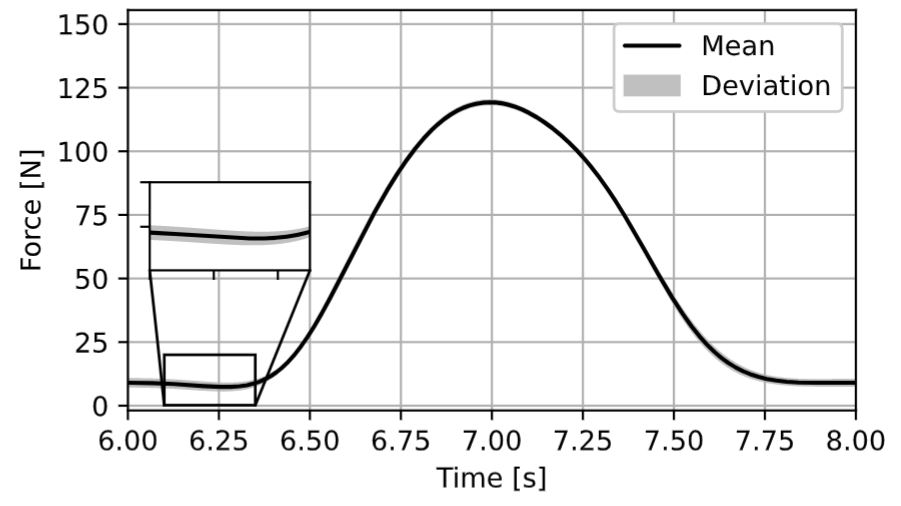}
            \caption{Force}
            \label{fig:hyd_force}
        \end{subfigure}
        \vspace{0.3cm}
    \caption{Hydraulic actuation system.}
    \label{fig:hydraulic}
\end{figure}

\begin{figure}[htb]
    \centering
        \begin{subfigure}[b]{0.45\columnwidth}
            \centering
            \includegraphics[width =1\columnwidth]{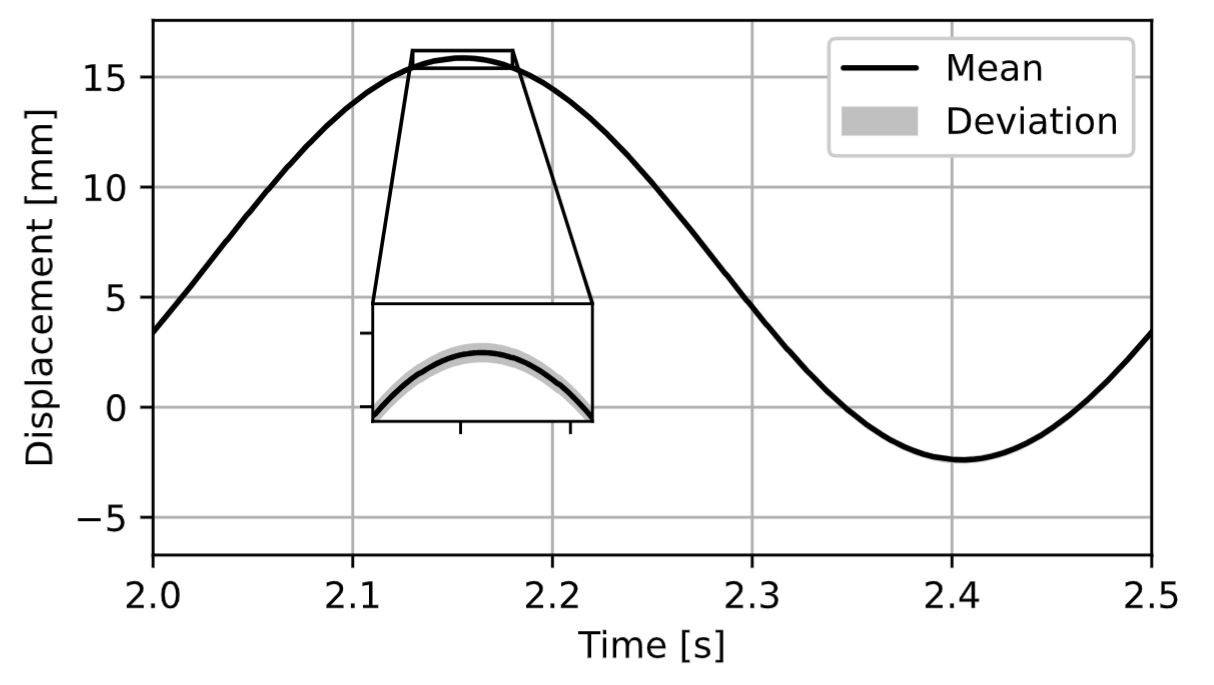}
            % [width =0.7\columnwidth]
            \caption{Displacement}
            \label{fig:elec_disp}
        \end{subfigure}
        %\hfill
        \begin{subfigure}[b]{0.45\columnwidth}
            \centering
            \includegraphics[width =1\columnwidth]{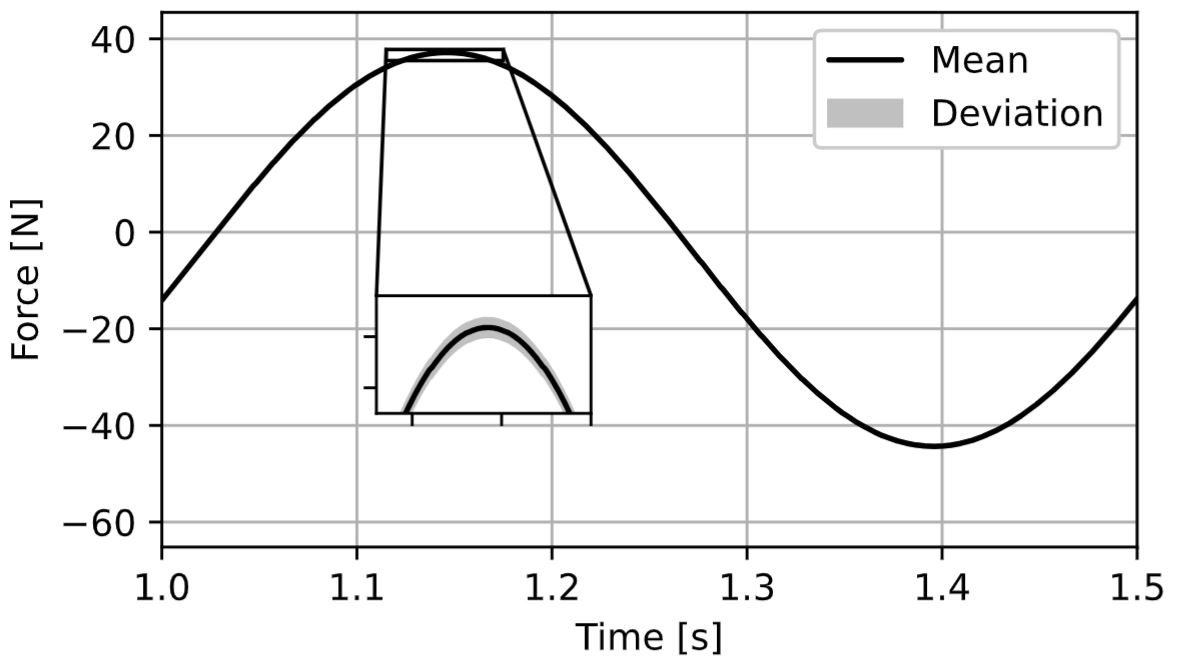}
            \caption{Force}
            \label{fig:elec_force}
        \end{subfigure}
        \vspace{0.3cm}
    \caption{Electric actuation system.}
    \label{fig:electric}
\end{figure}

The experimental setup for validation included proportional-integral (PI) force control on both linear actuators. We used the Moog E024 servovalve with \SI{10}{\mega\pascal} supply pressure, proportional gain $K_p = 1.5$, and integrative gain $K_i=1.5$ for the hydraulic actuator. The mechanical setup in this case was similar to what is present in Fig. \ref{fig:global}, without only the load car. Then, the reference force signal was a sine with \SI{50}{\newton} amplitude and \SI{0.1}{\hertz} frequency. One of the oscillations of the responses obtained from the encoder and load cells are in Fig.~\ref{fig:hydraulic}. We performed the same procedure for the electric actuator; however, the proportional gain was $K_p = 0.73$, the integrative was $K_i=0.03$, and the sine reference force amplitude was \SI{75}{\newton} and frequency \SI{2.0}{\hertz}. One of the oscillations of the responses obtained are in Fig.~\ref{fig:electric}. Each actuator had run the described setup ten times. Those experiments qualify the bench's reproducibility for the hydraulic and electric actuators based on their displacement and forces. As one can see by the displacement (Fig.~\ref{fig:hyd_disp}) and the force (Fig.~\ref{fig:hyd_force}), the deviation and mean obtained from ten experiments were very similar, pointing out a valid representation of the actual system, qualitatively and quantitatively.

\begin{figure}[htb]
    \centering
    \includegraphics[width =0.5\columnwidth]{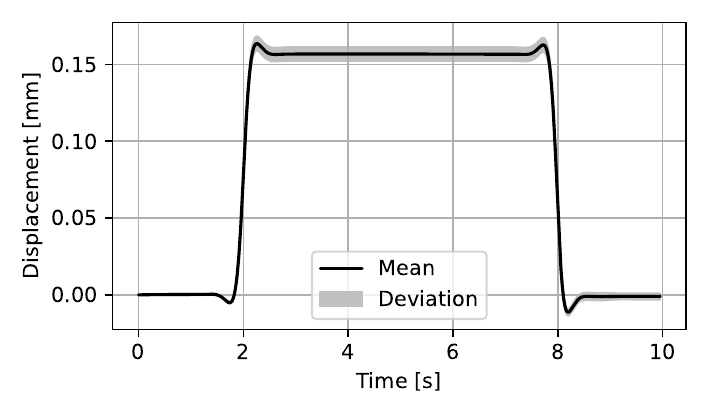}
    \caption{Displacemnt mesured at the blocked set up.}
    \label{fig:blocked}
\end{figure}

Using the electric actuator at its maximum operating force for a reference step signal, the displacement measurement when the actuator is blocked was approximate \SI{0.15}{\milli\meter} (Fig.\ref{fig:blocked}), the test was also repeated ten times. We can verify that the displacement comes mostly from the deformation of the load cell, therefore, it can be inferred that this displacement is irrelevant, verifying the objective of an assembly with very little backlash.

\begin{figure}[htb]
    \centering
    \includegraphics[width =0.55\columnwidth]{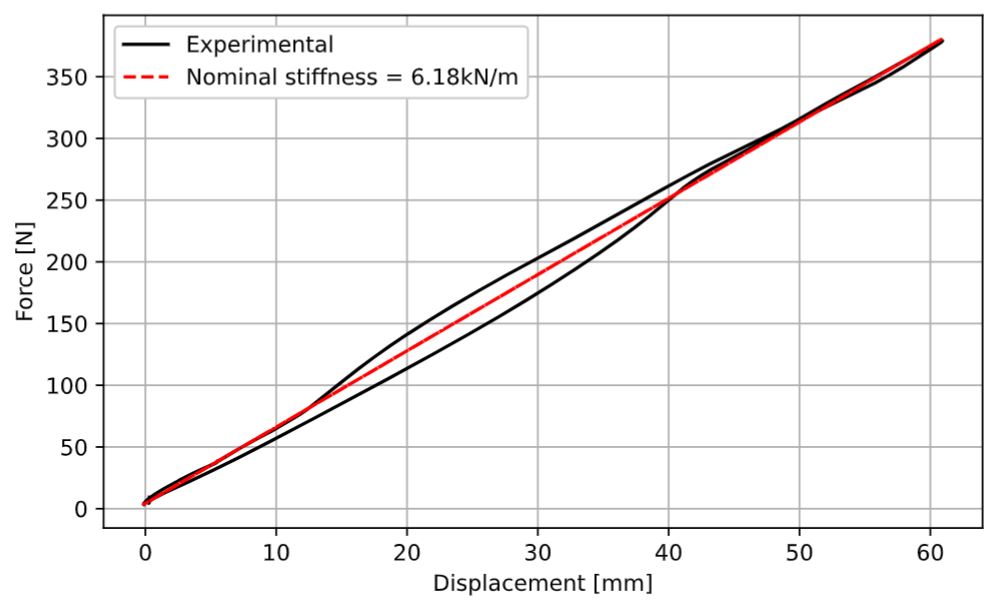}
    \caption{Linear regression of the experimental spring stiffness.}
    \label{fig:mola}
\end{figure}

In the third set of experiments, to assess frictional effects and sensor calibration, we manually compressed the spring with one end blocked while on the other end were the linear encoder and the force sensor. Force versus displacement experimental curves is in Fig.~\ref{fig:mola}. 

Performing a linear regression with the mean of ten experiments, with a P correlation coefficient of $R=0.9987$, the identified stiffness of the spring was \SI{6.18}{\kilo\newton}. The expected value was \SI{6}{\kilo\newton}. This variation was the result of the friction that exists between the sliders and the guide \citep{Zanette23}, and it is still possible to see the effect of viscous friction more present in the transition between ends where the movement velocity was more expressive.

%%%%%%%%%%%%%%%%%%%%%%%%%%%%%%%%%%%%%%%%%%%%%%%%%%%%%%%%%%%%%%

\section{CONCLUSION AND NEXT STEPS}

The bench has proven to be robust and capable of meeting project requirements. The IC2D is a dependable and adaptable engineering platform for designing and testing control methods using software, particularly force and impedance/admittance controllers. Currently, the bench works with linear actuators despite supporting the use of rotational motors, provided they have a high level of torque to ensure the effective movement of the mobile parts. Moreover, it is imperative to meticulously check the backlashes in the nuts and spindles, as they are typically used for position control and not force control scenarios. The next step is to model and validate the bench actuation systems, which are already in progress and undergoing validation. Modeling is fundamental for the optimal performance of model-based control design. Besides, a more comprehensive study of the friction in the bench sliders and a more intricate model proposal is ongoing.

\addtolength{\textheight}{-12cm}   % This command serves to balance the column lengths

\section{ACKNOWLEDGEMENTS}

The authors would like to thank the Legged Robotics Group of the Robotics Laboratory of the São Carlos School of Engineering, University of São Paulo and the \textit{Equitron Automação}, in special Prof. Dr. Eng. José Guilherme Sabe, for all the devices and consultancy. This work is supported by São Paulo Research Foundation (FAPESP) under grants 2018/15472-9, 2021/03373-9, and 2021/09244-6 and CAPES 88887.817139/2023-00.

\section{REFERENCES} \label{Sec:references}

\bibliographystyle{abcm}
\renewcommand{\refname}{}
\bibliography{bibfile}

\section{RESPONSIBILITY NOTICE}
The authors are solely responsible for the printed material included in this paper.

\end{document}